\documentclass[daat]{ipart}
\usepackage{tabularx}
\RequirePackage{hyperref}
\usepackage{placeins}
\startlocaldefs
\theoremstyle{plain}

\endlocaldefs
\startlocaldefs
\theoremstyle{plain}

\endlocaldefs

\pubyear{2025}
\volume{1}
\issue{1}
\firstpage{1}
\lastpage{4}
\arxiv{0000.0000}

\usepackage{pdfpages}
\usepackage{latexsym}
\usepackage{amssymb}
\usepackage{amsmath}
\usepackage{algorithm, algorithmic}
\usepackage{xcolor}
\usepackage{makecell}
\usepackage{multirow}
\usepackage{adjustbox}
\usepackage{booktabs}
\usepackage{subcaption}
\usepackage{wrapfig}
\usepackage[most]{tcolorbox}  
\tcbuselibrary{skins,breakable}

\newtcolorbox{promptbox}[1][]{
    enhanced,
    breakable,
    colback=yellow!5,          
    colframe=yellow!50!orange, 
    colbacktitle=yellow!15,    
    coltitle=black,            
    title={#1},
    fonttitle=\bfseries\ttfamily,
    boxrule=0.6pt,
    arc=3pt,
    outer arc=3pt,
    left=8pt, right=8pt,
    top=6pt, bottom=6pt,
}
\tcbuselibrary{skins, breakable}

\usepackage{graphicx}

\begin{document}

\begin{frontmatter}

\title[A sample document]{Semantic Distance Measurement based on Multi-Kernel Gaussian Processes}

\begin{aug}
    \author{\inits{Y.}\fnms{Yinzhu} 
    \snm{Cheng}\ead[label=e1]{chengyinzhu@bimsa.cn}},
    \address{Institute of Statistics and Big Data, Renmin University of China\\
    Beijing 100872, P.R. China\\
    Beijing Institute of Mathematical Sciences and Applications\\
    Beijing 101408, P.R. China\\
    \printead{e1}}
    \author{\inits{H.}\fnms{Haihua} 
    \snm{Xie}\thanksref{t2}\ead[label=e2]{haihuaxie@bimsa.cn}},
    \address{Beijing Institute of Mathematical Sciences and Applications\\
    Beijing 101408, P.R. China\\
    \printead{e2}}    
    \author{\inits{Y.}\fnms{Yaqing} 
    \snm{Wang}\ead[label=e3]{wangyaqing@bimsa.cn}},
    \address{Beijing Institute of Mathematical Sciences and Applications\\
    Beijing 101408, P.R. China\\
    \printead{e3}}
    \author{\inits{M.}\fnms{Miao} 
    \snm{He}\ead[label=e4]{hemiao@bimsa.cn}},
    \address{Beijing Institute of Mathematical Sciences and Applications\\
    Beijing 101408, P.R. China\\
    \printead{e4}}
    \author{\inits{M.}\fnms{Mingming} 
    \snm{Sun}\thanksref{t2}\ead[label=e5]{sunmingming@bimsa.cn}}
    \address{Beijing Institute of Mathematical Sciences and Applications\\
    Beijing 101408, P.R. China\\
    \printead{e5}}
    \thankstext{t2}{Corresponding authors.}
\end{aug}

\received{\sday{3} \smonth{1} \syear{2022}}

\begin{abstract}
Semantic distance measurement is a fundamental problem in computational linguistics, providing a quantitative characterization of similarity or relatedness between text segments, and underpinning tasks such as text retrieval and text classification. From a mathematical perspective, a semantic distance can be viewed as a metric defined on a space of texts or on a representation space derived from them. However, most classical semantic distance methods are essentially fixed, making them difficult to adapt to specific data distributions and task requirements. In this paper, a semantic distance measure based on multi-kernel Gaussian processes (MK-GP) was proposed. The latent semantic function associated with texts was modeled as a Gaussian process, with its covariance function given by a combined kernel combining Mat\'ern and polynomial components. The kernel parameters were learned automatically from data under supervision, rather than being hand-crafted. This semantic distance was instantiated and evaluated in the context of fine-grained sentiment classification with large language models under an in-context learning (ICL) setup. The experimental results demonstrated the effectiveness of the proposed measure.

	\end{abstract} 


\end{frontmatter}

\section{Introduction}

Measuring semantic distance, understood as how similar or related two texts are in meaning, is a fundamental problem in computational linguistics~\citep{BudanitskyHirst2006}. Effective distance measures support information retrieval~\citep{ManningIR2008}, text classification~\citep{Sebastiani2002}, and recent large language model (LLM) paradigms such as in-context learning and retrieval-augmented generation~\citep{Brown2020}.

Classical approaches can be roughly grouped into three families: (1) \textbf{Statistical methods} in the raw text space, such as TF--IDF and BM25~\citep{ManningIR2008}; (2) \textbf{Embedding-based methods}, which obtain dense vectors from neural encoders such as BERT and Sentence-BERT~\citep{ReimersGurevych2019SBERT} and then apply simple metrics like cosine similarity; and (3) \textbf{Kernel-based methods}, which define similarity via positive-definite kernels in an implicit feature space~\citep{ScholkopfSmola2002}.

Despite substantial progress, learning a representation and metric that encode task-specific semantic relations remained challenging when semantics were subtle, context-dependent, or organized along fine-grained categories. Multi-class sentiment classification was a prime example: labels such as ``very positive'' and ``slightly positive'' lay close on an affective continuum yet corresponded to distinct attitudes~\citep{SocherEtAl2013}. Texts in adjacent sentiment categories often shared similar lexical patterns, making them hard to separate using frequency-based methods that relied on term occurrence and inverse document frequency~\citep{Sebastiani2002}. Embedding-based methods with cosine similarity further assumed a globally smooth, approximately linear geometry of semantic space, which blurred fine distinctions and underrepresented non-linear phenomena such as negation, intensifiers, and sarcasm~\citep{Gutierrez2016Ordinal}.

This paper addresses these limitations by proposing a semantic distance framework based on multi-kernel Gaussian processes (MK-GP). The unknown task-specific function that maps text representations to sentiment values is treated as being drawn from a Gaussian process prior with constant mean and a combined covariance function. Within this framework, semantic similarity is governed by the geometry implied by the covariance function rather than by a fixed embedding space with a hand-designed metric. Two texts are considered close when the Gaussian process prior expects similar latent outputs for them; conversely, they are far apart when the prior allows the function to vary substantially between the corresponding points. This induces a natural notion of task-aware distance linked to how uncertain the model is about interpolating between texts, providing a principled way to express confidence in fine-grained distinctions when data are scarce or ambiguous.

This paper makes the following contributions:
\begin{enumerate}
    \item \textbf{MK-GP-based semantic distance framework.}  
    A Gaussian-process–based semantic distance framework is proposed, in which a combined covariance function defines a learned, task-specific covariance geometry over representations and induces an explicit distance measure, replacing fixed vector-space metrics.

    \item \textbf{MK-GP learning algorithm.}  
    A multi-kernel Gaussian process construction is designed that combines Mat\'ern and polynomial kernels, and a training procedure is developed to learn the combined covariance and its hyperparameters from data. The learned MK-GP geometry yields a task-specific semantic distance that is compatible with modern LLM-based feature extractors and retrieval frameworks.

    \item \textbf{Fine-grained sentiment ICL instantiation and analysis.}  
    The framework is instantiated in the setting of in-context learning for fine-grained sentiment tasks, using the learned MK-GP semantic distance to select support examples for LLM prompts. The analysis shows how the induced geometry captures subtle sentiment differences and ordinal structure along the sentiment continuum, and how MK-GP–based example selection impacts downstream ICL performance.
\end{enumerate}

The remainder of this paper is organized as follows. Section~\ref{sec:related} reviews related work on semantic distance metrics, kernel-based approaches for textual semantic modeling, and semantic distance measurement in the era of LLMs. Section~\ref{sec:method} presents the proposed MK-GP methodology, detailing the multi-kernel Gaussian process formulation for semantic distance estimation and its integration with in-context learning. Section~\ref{sec:experiments} reports the experiments and analyzes the empirical results. Finally, Section~\ref{sec:conclusion} concludes the paper and discusses directions for future work.
	
	\section{Related Work}
\label{sec:related}
\subsection{Semantic Distance Metrics} 

Text distance measures aim to quantify the similarity between pieces of text (e.g., sentences or documents). A smaller distance indicates a higher similarity. Three classical families of text distance/similarity measures are now introduced. 

\subsubsection{Statistical methods in the raw text space.}
Classical information retrieval models such as TF--IDF and BM25 measure similarity via word overlap and term frequency statistics, often within a probabilistic relevance framework~\citep{SaltonBuckley1988_RW}. These methods work well when lexical overlap correlates with meaning, but fail on subtle, contextual phenomena such as negation or pragmatic implicature that require modeling beyond surface forms.

\subsubsection{Embedding-based similarity.}
Dense word and sentence embeddings support distance measures such as cosine similarity. Canonical examples include neural word embeddings and transformer-based encoders such as BERT and Sentence-BERT~\citep{MikolovEtAl2013_RW}. These methods usually assume a globally smooth, approximately linear geometry in the learned embedding space, which is often inadequate for structured, ordered semantic spaces such as fine-grained sentiment scales and continuous affective dimensions~\citep{PreotiucPietroEtAl2016_RW}.

\subsubsection{Kernel methods.}
Kernel methods instantiate similarity via a positive-definite kernel
\(k(x_i, x_j)\), which can be viewed as an inner product in an implicit
feature space through a feature map \(\varphi\), thereby supporting rich non-linear geometries in a reproducing kernel
Hilbert space~\citep{ScholkopfSmola2002_RW}. For text, string and tree
kernels define similarity by counting shared subsequences or subtrees,
and have been used for tasks such as document classification and
syntactic parsing. In most of these applications, the functional form of
the kernel and its hyperparameters (e.g., length-scales, polynomial
degrees, or mixture weights) are largely specified by hand or selected
via heuristic search and cross-validation, often requiring substantial
tuning expertise. Moreover, the kernel is typically treated as a
convenient component of a predictor, rather than as an explicitly
calibrated semantic metric tailored to fine-grained sentiment.

\subsubsection{Our approach.}
Texts are modeled with a Gaussian process, and its
covariance function is taken as the primary semantic similarity object. 
A combined Mat\'ern + polynomial kernel is designed whose geometry is intended to
capture both local smoothness and global ordinal trends in sentiment, and its hyperparameters are learned from data by maximizing the GP marginal
likelihood. This hyperparameter learning automatically
balances data fit and model complexity, yielding a kernel that is
adapted to the observed sentiment annotations rather than hand-tuned.

\subsection{Applications of Semantic Distance Measurement in LLM-based Systems}

In modern large language model (LLM) pipelines it is an  systems component
that decides which pieces of text an LLM will ever see at
inference time. Most workflows first map texts to continuous embeddings
using neural encoders (e.g., BERT-style encoders or encoder-only
embedding models such as Sentence-BERT)~\citep{DevlinEtAl2019_RW},
and then endow the embedding space with a simple, globally defined
metric such as Euclidean distance or cosine similarity. This
“encode-then-distance” template is attractive because it is easy to
scale, but it implicitly assumes that a single,
task-agnostic metric on a fixed embedding space is sufficient to capture
all relevant semantic relationships. In particular, it offers no
explicit control over which semantic dimensions (e.g., topic, style,
sentiment) are prioritized, nor any notion of uncertainty about
similarity in regions poorly covered by training data.

In in-context learning (ICL), semantic distance is used to select relevant examples from the context. This involves choosing the most similar examples based on the distance between the query and context texts in the embedding space. By prioritizing label-compatible and semantically similar instances, ICL adapts the model's response to the specific context of the task. This approach leverages similarity to improve the relevance of the selected examples, guiding the model's inference in a few-shot setting~\citep{BrownEtAl2020_RW}.

In retrieval-augmented generation (RAG), semantic distance is used to retrieve relevant documents from an external corpus based on query-document similarity in the embedding space~\citep{KarpukhinEtAl2020_RW_DPR}. Unlike ICL, which relies solely on in-context examples, RAG enhances generation by retrieving external knowledge to provide additional context. Both methods depend on the quality of the semantic distance metric, as it directly affects the relevance of the selected examples or retrieved documents.

\section{Methodology}
\label{sec:method}
\begin{figure*}[t]
    \centering
    \includegraphics[width=\textwidth]{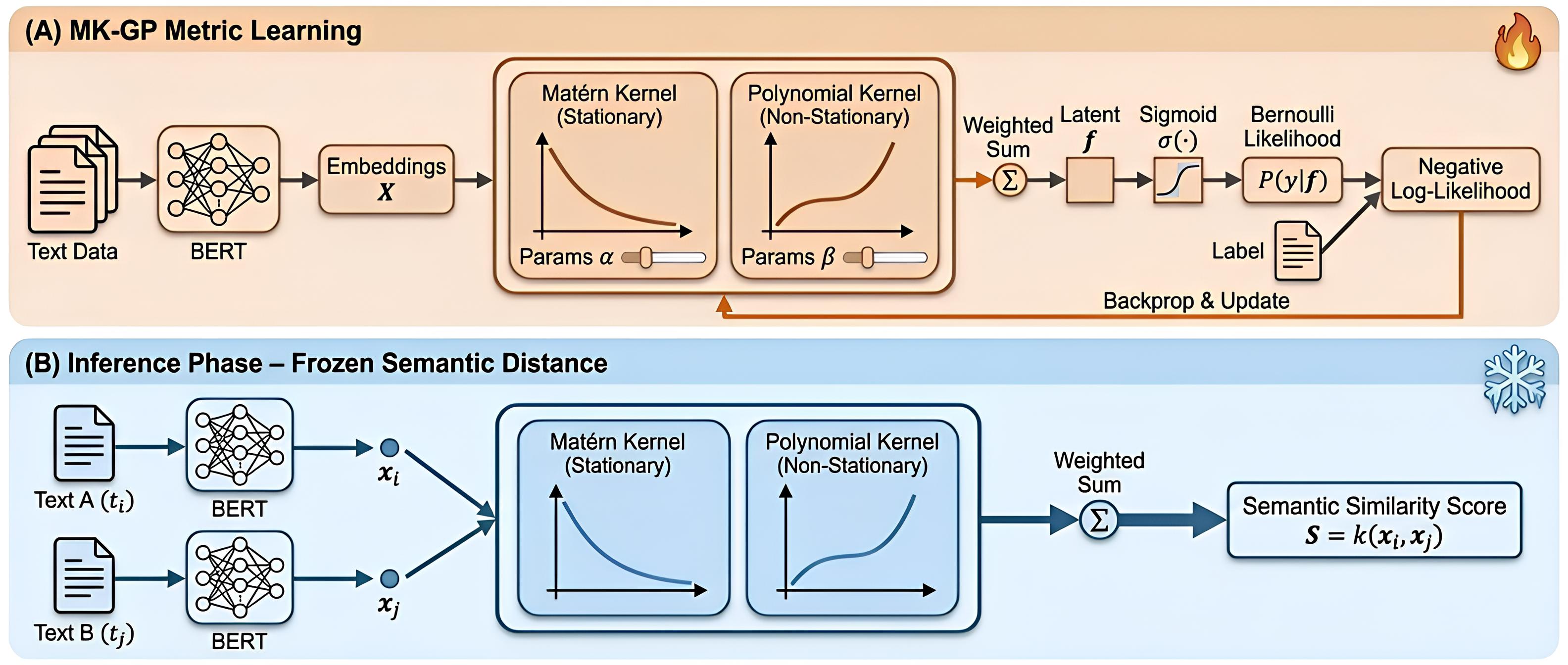}
    \caption{
        Overall workflow of the proposed MK-GP framework.
        (A) MK-GP metric learning: text embeddings are obtained from a pretrained encoder,
        followed by multi-kernel Gaussian process training with Mat\'ern and polynomial kernels,
        whose parameters are updated via negative log-likelihood minimization.
        (B) Inference phase: the learned kernel parameters are frozen to compute the
        MK-GP semantic similarity score \(k(\boldsymbol{x}_i, \boldsymbol{x}_j)\)
        between text embeddings.
    }
    \label{fig:mk-gp-pipeline}
\end{figure*}

This section first introduces how to learn a task-aware semantic distance over the embedding space using a multi-kernel Gaussian process (MK-GP) (in subsection \ref{sec:metric}). Kernel-based models are particularly attractive in small-sample regimes because their capacity is controlled by the choice of kernel and a few hyperparameters, enabling robust generalization from limited data~\citep{rasmussen2006gaussian}. Recent work on further shows that lightweight kernel combinations can be especially effective for few-shot classification~\citep{zhang2021taskadaptive}. Building on this MK-GP-based semantic distance, the MK-GP-based ICL framework is then described in subsection \ref{sec:rdmcsa}, integrating the learned metric with class-level rationales and in-context prompting for multi-class sentiment analysis.

To conclude, the MK-GP model first learns a task-aware semantic distance over
the embedding space via supervised kernel parameter estimation, and then reuses
the learned kernel as a frozen semantic similarity measure at inference time.
The overall workflow of the proposed MK-GP framework, including both the metric
learning stage and the frozen inference stage, is illustrated in
Figure~\ref{fig:mk-gp-pipeline}.

\subsection{Task-Aware Semantic Distance with Multi-Kernel Gaussian Processes}
\label{sec:metric}

\begin{figure*}[t]
  \centering
  \includegraphics[
    width=\linewidth,
    trim=0 70em 0 50em, 
    clip
  ]{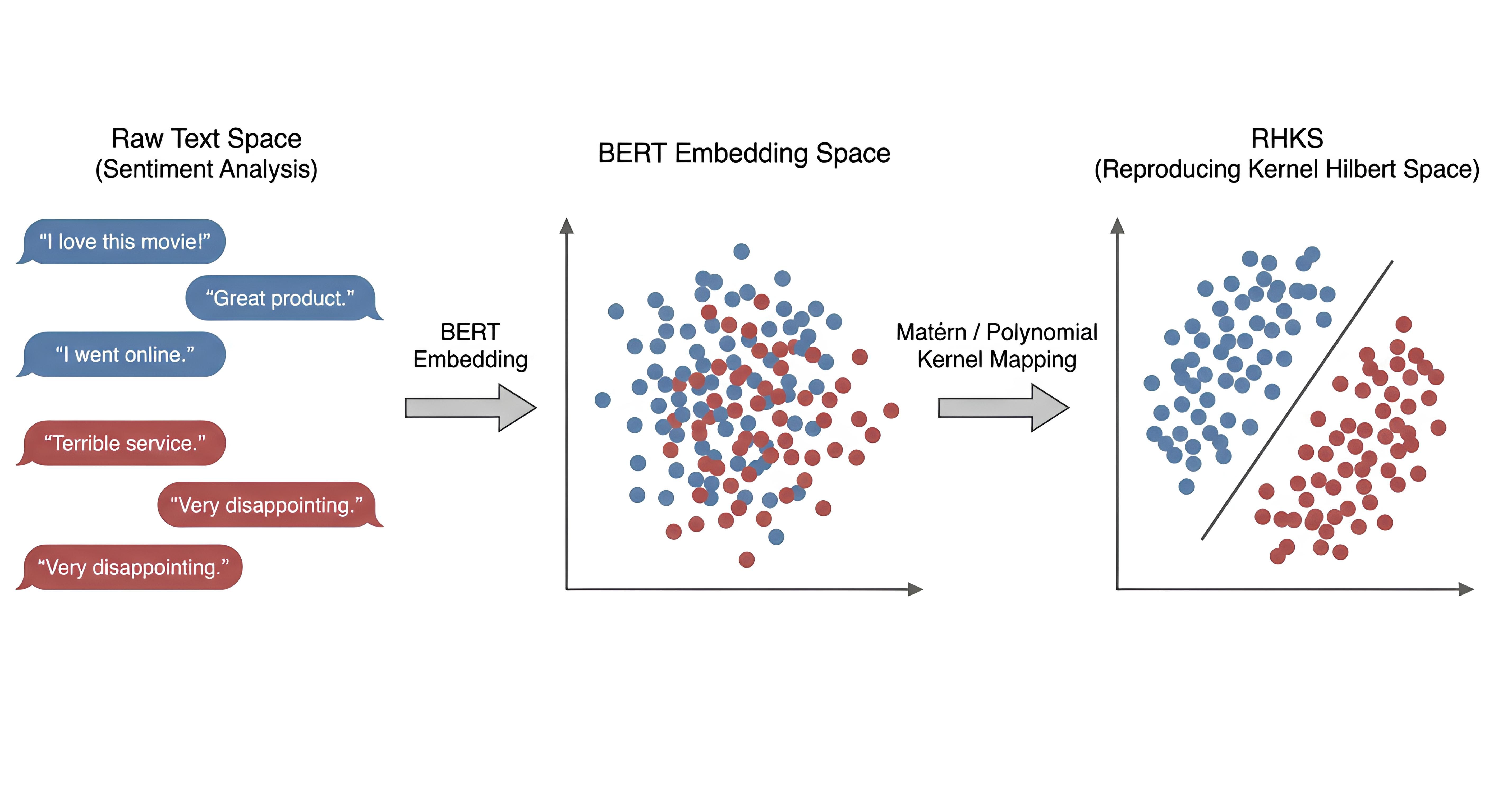}
  \caption{Illustration of the mapping from original texts to the embedding space and then to the kernel-induced Hilbert space.}
  \label{fig:embedding}
\end{figure*}

Given an annotated text dataset for a classification task, \( \mathcal{D} = \{(t_i, y_i)\}_{i=1}^K \), where \( y_i \) denotes the class label of the text \( t_i \). Each text input \( t_i \) is mapped to a vector representation \( \boldsymbol{x}_i \in \mathbb{R}^d \) using a sentence encoder.
The objective is to learn a semantic similarity function
\( k(\boldsymbol{x}_i, \boldsymbol{x}_j) \) that is consistent with the label structure of the task: representations belonging to the same class should be close in the embedding space, whereas those from different classes should be well separated, as illustrated in Figure~\ref{fig:embedding}.

\subsubsection{The Multi-Kernel Gaussian Processes}

This similarity is modeled with a Gaussian process (GP) over the embedding space~\cite{wang2023intuitive}.
For a multi-class problem with \(u\) labels,  a latent scalar-valued function is introduced.
\begin{equation}
    f_c : \mathbb{R}^d \to \mathbb{R}, \qquad c \in \{1,\dots,u\},
\end{equation}
for each class, and collect their pointwise evaluations as
\begin{equation}
    \boldsymbol{f}(\boldsymbol{x})
    = [f_1(\boldsymbol{x}), \dots, f_u(\boldsymbol{x})]^\top \in \mathbb{R}^u,
\end{equation}
so that at any embedding \(\boldsymbol{x}\), the vector \(\boldsymbol{f}(\boldsymbol{x})\)
contains one latent score per class.

Each class-specific latent function is assigned an independent GP prior:
\begin{equation}
    f_c(\boldsymbol{x})
    \sim \mathcal{GP}\big(e_c(\boldsymbol{x}),\, k(\boldsymbol{x}, \boldsymbol{x}')\big),
    \qquad c = 1,\dots,u,
\end{equation}
where \(e_c(\cdot)\) is the mean function for class \(c\) and
\(k(\cdot,\cdot)\) is a covariance function (kernel) shared across classes. Let \(\boldsymbol{X} = \{\boldsymbol{x}_i\}_{i=1}^K\) denote the training
embeddings.
For a fixed class \(c\), the latent function values are stacked on these inputs into
\begin{equation}
    \boldsymbol{f}_c
    = [f_c(\boldsymbol{x}_1), \dots, f_c(\boldsymbol{x}_K)]^\top
    \in \mathbb{R}^K.
\end{equation}
By the definition of a Gaussian process, the joint distribution of
\(\boldsymbol{f}_c\) at any finite collection of inputs is multivariate normal:
\begin{equation}
    p(\boldsymbol{f}_c \mid \boldsymbol{X})
    = \mathcal{N}\big(\boldsymbol{m}_{\boldsymbol{X},c},\,
                      \boldsymbol{K}_{\boldsymbol{X}\boldsymbol{X}}\big),
\end{equation}
where the mean vector
\(\boldsymbol{m}_{\boldsymbol{X},c} \in \mathbb{R}^K\) has entries
\(\,[\boldsymbol{m}_{\boldsymbol{X},c}]_i = e_c(\boldsymbol{x}_i) = \mu_c\),
and the Gram matrix
\(\boldsymbol{K}_{\boldsymbol{X}\boldsymbol{X}} \in \mathbb{R}^{K \times K}\)
is shared across classes with entries
\([\boldsymbol{K}_{\boldsymbol{X}\boldsymbol{X}}]_{ij} = k(\boldsymbol{x}_i, \boldsymbol{x}_j)\).

Assuming the GPs \(\{f_c\}_{c=1}^u\) are a priori independent, the joint prior
over all latent values simply factorizes over classes as:
\begin{equation}
    p(\boldsymbol{f} \mid \boldsymbol{X})
    = \prod_{c=1}^u p(\boldsymbol{f}_c \mid \boldsymbol{X}),
\end{equation}
where \(\boldsymbol{f}\) denotes the collection of all class-wise latent values.

The likelihood term \(p(\boldsymbol{y} \mid \boldsymbol{f})\) links the latent functions to the observed labels.
Each class \(c\) is associated with a latent function value \(f_c(\boldsymbol{x}_i)\), which is transformed into a probability via a logistic (sigmoid) link and a Bernoulli likelihood.
Let \(y_i^{(c)} \in \{0,1\}\) denote the indicator of whether sample \(i\) belongs to class \(c\).
For input \(\boldsymbol{x}_i\) and class \(c\), the conditional distribution is
\begin{equation}
	p\!\left(y_i^{(c)} \mid f_c(\boldsymbol{x}_i)\right) 
	= \mathrm{Ber}\!\Big(y_i^{(c)};\, \sigma(f_c(\boldsymbol{x}_i))\Big),
\end{equation}
with
\begin{equation}
	\sigma(t) = \frac{1}{1+e^{-t}}.
\end{equation}
Assuming conditional independence across data points and classes, the overall likelihood factorizes as
\begin{equation}
	\label{eq:bern_lik}
	p(\boldsymbol{y}\mid \boldsymbol{f}) 
	= \prod_{c=1}^u \prod_{i=1}^K p\!\left(y_i^{(c)} \mid f_c(\boldsymbol{x}_i)\right).
\end{equation}

Let \( \boldsymbol{f} \) denote all latent values and \( \boldsymbol{y} \) the observed labels.
The kernel parameters (including base-kernel hyperparameters and mixing weights) are estimated by minimizing the negative log marginal likelihood~\cite{artemev2021tighter}:
\begin{equation}
	\mathcal{L}
	= - \log \int p(\boldsymbol{y} \mid \boldsymbol{f})\,
	p(\boldsymbol{f} \mid \boldsymbol{X})\,
	d\boldsymbol{f},
\end{equation}

\subsubsection{Attributes of the Kernels}

The kernel function \( k \) is parameterized as a weighted combination of Mat\'ern and polynomial kernels:
\begin{equation}
	\label{eq:mk}
	\begin{aligned}
		k(\boldsymbol{x}_i, \boldsymbol{x}_j)
		= & \sum_{n=1}^N \alpha_{n}\, k_{\text{Mat\'ern},n}(\boldsymbol{x}_i, \boldsymbol{x}_j) \\
		& + \sum_{m=1}^M \beta_{m}\, k_{\text{Poly},m}(\boldsymbol{x}_i, \boldsymbol{x}_j),
	\end{aligned}
\end{equation}
where all base-kernel hyperparameters and mixing weights
\( \alpha_n, \beta_m \ge 0 \) are learned from data.
This multi-kernel construction allows the Gaussian process to combine complementary local and global structures in the embedding space~\cite{ghasempour2023multiple}.

Each polynomial kernel~\cite{song2021fast} has the form
\begin{equation}
	k_{\text{Poly},m}(\boldsymbol{x}_i, \boldsymbol{x}_j)
	= \big(\gamma_m \, \langle \boldsymbol{x}_i, \boldsymbol{x}_j \rangle + c_m\big)^{d_m},
\end{equation}
where \( \gamma_m \) is a scaling factor, \( c_m \) is an offset (both learnable), \( d_m \) is the polynomial degree (treated as a hyperparameter), and
\( \langle \boldsymbol{x}_i, \boldsymbol{x}_j \rangle \) denotes the standard inner product.
Polynomial kernels capture global, potentially non-linear interactions between text embeddings and can emphasize margin-like structures in the induced feature space.

Each Mat\'ern kernel~\cite{porcu2024matern} is defined as
\begin{equation}
	k_{\text{Mat\'ern},n}(\boldsymbol{x}_i, \boldsymbol{x}_j)
	= \sigma_n^2 \,
	\frac{2^{1-\nu_n}}{\Gamma(\nu_n)}
	\Bigg( \sqrt{2\nu_n}\,\frac{r_{ij}}{\ell_n} \Bigg)^{\nu_n}
	K_{\nu_n}\!\Bigg( \sqrt{2\nu_n}\,\frac{r_{ij}}{\ell_n} \Bigg),
\end{equation}
where \( r_{ij} = \|\boldsymbol{x}_i - \boldsymbol{x}_j\|_2 \),
\( \sigma_n^2 \) is a signal variance,
\( \ell_n \) is a length-scale,
\( \nu_n > 0 \) controls smoothness,
\( \Gamma(\cdot) \) is the Gamma function,
and \( K_{\nu}(\cdot) \) is the modified Bessel function of the second kind.
The function \( K_{\nu}(z) \) can be expressed in terms of the modified Bessel functions of the first kind \( I_{\nu}(z) \) as
\begin{equation}
	K_{\nu}(z)
	= \frac{\pi}{2} \,
	\frac{I_{-\nu}(z) - I_{\nu}(z)}{\sin(\nu \pi)},
\end{equation}
where \( I_{\nu}(z) \) admits the series representation
\begin{equation}
	I_{\nu}(z)
	= \sum_{k=0}^{\infty}
	\frac{\big(\frac{z}{2}\big)^{\nu + 2k}}{k!\,\Gamma(\nu + k + 1)}.
\end{equation}
By mixing multiple Mat\'ern and polynomial components with different scales and degrees, the MK-GP learns a task-adaptive semantic metric that can better align with the complex geometry of multi-class labels.

\subsubsection{ MK-GP Used for Semantic Distance Measurement}

By Mercer's theorem~\cite{thickstun2019mercer}, there exists a Hilbert space \( \mathcal{H} \) and a feature map
\( \phi: \mathcal{X} \to \mathcal{H} \) such that
\begin{equation}
	k(\boldsymbol{x}_i, \boldsymbol{x}_j)
	= \langle \phi(\boldsymbol{x}_i), \phi(\boldsymbol{x}_j) \rangle_{\mathcal{H}}.
\end{equation}
This induces the squared distance
\begin{align}
	\big\|\phi(\boldsymbol{x}_i) - \phi(\boldsymbol{x}_j)\big\|^2
	&= k(\boldsymbol{x}_i, \boldsymbol{x}_i)
	- 2 k(\boldsymbol{x}_i, \boldsymbol{x}_j)
	+ k(\boldsymbol{x}_j, \boldsymbol{x}_j).
\end{align}
After normalizing embeddings, the self-kernel terms are approximately constant, so a larger kernel value corresponds to a smaller kernel-induced distance.
The task-aware semantic similarity between texts \( t_i \) and \( t_j \) is therefore defined as
\begin{equation}
	\mathrm{sim}(t_i, t_j) = k(\boldsymbol{x}_i, \boldsymbol{x}_j),
\end{equation}

In this paper, this kernel-induced similarity is primarily used to drive in-context demonstration selection in the MK-GP-based ICL framework, but in principle the same task-aware distance can also support topic discovery (e.g., clustering in the kernel space) and retrieval-augmented generation by replacing task-agnostic metrics such as BM25 or cosine similarity in the retriever.

\subsection{In-Context Learning with MK-GP}
\label{sec:rdmcsa}

Building on the task-aware semantic distance learned by the MK-GP model \( \mathcal{G} \) in Section~\ref{sec:metric}, the MK-GP-based ICL framework instantiates a multi-class sentiment classifier in an in-context learning (ICL) paradigm. Concretely, it combines (i) class-level classification rationales induced from a small, label-balanced coreset and (ii) MK-GP-based similarity-driven demonstration selection, yielding prompts that are both geometrically and linguistically aligned with the label structure of the MCSA task.

\subsubsection{Balanced Coreset for Rationale Induction}
\label{Coreset}

To obtain high-quality class-level rationales, the MK-GP-based ICL framework follows the balanced coreset construction strategy of prior work~\cite{xie2025rd} to build a small yet informative subset \( \mathcal{B} \subset \mathcal{D} \). For each labeled text \((t_i, y_i) \in \mathcal{D}\), let \(\boldsymbol{x}_i \in \mathbb{R}^d\) denote its embedding, and let \(\mathcal{D}_c \subset \mathcal{D}\) be the set of samples with label \(c\). The centroid of class \(c\) in the embedding space is computed as
\begin{equation}
	\boldsymbol{\mu}_c = \frac{1}{|\mathcal{D}_c|} \sum_{j : y_j = c} \boldsymbol{x}_j.
\end{equation}
An importance weight is then assigned to each sample according to its squared Euclidean distance from the corresponding class centroid,
\begin{equation}
	w(t_i, y_i) = \big\|\boldsymbol{x}_i - \boldsymbol{\mu}_{y_i}\big\|_2^2,
\end{equation}
so that examples that are farther from the centroid---and thus more likely to lie near class boundaries or to be atypical but informative---receive larger weights. Within each class, these importance weights are normalized to form a sampling distribution
\begin{equation}
	P_c(t_i) = \frac{w(t_i, y_i)}{\sum_{j : y_j = c} w(t_j, y_j)}.
\end{equation}

Stratified importance-weighted sampling is then performed independently for each class \(c\). If \(|\mathcal{D}_c| \le \lambda_{\mathcal{B}}\), all samples from class \(c\) are included in the coreset. Otherwise, a subset \(\mathcal{B}_c \subset \mathcal{D}_c\) of size \(|\mathcal{B}_c| = \lambda_{\mathcal{B}}\) is drawn from \(\mathcal{D}_c\) according to the sampling probabilities \(P_c(t_i)\). The final balanced coreset is obtained by aggregating all class-wise subsets,
\begin{equation}
	\mathcal{B} = \bigcup_c \mathcal{B}_c.
\end{equation}
This procedure yields a small coreset that is approximately label-balanced while focusing on boundary-adjacent and atypical examples that provide good coverage of the semantic space and are particularly informative for fine-grained decision making.

Given \( \mathcal{B} \) and the list of sentiment labels, an LLM is then prompted to induce class-level classification rationales. For each label, the model summarizes the characteristic lexical patterns, semantic--pragmatic cues, and domain- or attribute-specific associations that distinguish it from the others (Figure~\ref{fig:crg-prompt}). The resulting rationales \( \mathcal{R} \) provide compact, human-readable descriptions of each sentiment class and are reused in downstream ICL prompts, helping align label semantics with the behavior of the LLM classifier~\cite{wang2025tutorialllmreasoningrelevant}.

\begin{figure}[H]
	\centering
	\begin{promptbox}[{\large\bfseries Classification Rationale Generation}]
		\small\ttfamily
		Based on the representative examples provided below, generate detailed
		descriptions for each sentiment label.
		
		\vspace{0.5em}
		\textbf{Examples:} \{Balanced Coreset $\mathcal{B}$\} \\
		\textbf{Sentiment Labels:} \{\textit{str(label\_list)}\}
		
		\vspace{0.5em}
		For each sentiment label, provide a comprehensive description covering:
		\begin{itemize}
			\item Lexical Patterns
			\item Semantic--Pragmatic Features
			\item Domain--Attribute Associations
		\end{itemize}
	\end{promptbox}
	\caption{Prompt template for generating class-level classification rationales
		from the balanced coreset $\mathcal{B}$.}
	\label{fig:crg-prompt}
\end{figure}

\subsubsection{MK-GP-Based Demonstration Selection and In-Context Prediction}
\label{ICL}

Let \( \mathcal{Y} = \{1,\dots,u\} \) denote the sentiment label set and
\( \mathcal{D} = \{(t_i, y_i)\}_{i=1}^K \) the annotated MCSA dataset, where
each text \( t_i \) is mapped to an embedding \( \boldsymbol{x}_i \in \mathbb{R}^d \)
by a sentence encoder.
For a new query text \( t_0 \) with embedding \( \boldsymbol{x}_0 \), the MK-GP-based ICL framework uses
the kernel function of the trained MK-GP model \( \mathcal{G} \) to compute a
task-aware similarity score
\begin{equation}
	s_i(t_0) = k(\boldsymbol{x}_0, \boldsymbol{x}_i),
	\quad i = 1,\dots,K.
\end{equation}
Texts with larger kernel values \( s_i(t_0) \) are considered more similar to \( t_0 \)
in the kernel-induced semantic space (cf.\ Section~\ref{sec:metric}).

Let \( I_S(t_0) \subset \{1,\dots,K\} \) be the index set of the top-\(S\) most similar
instances according to \( s_i(t_0) \), and let
\begin{equation}
	I_S(t_0) = \{i_1,\dots,i_S\}
	\quad \text{with} \quad
	s_{i_1}(t_0) \ge \dots \ge s_{i_S}(t_0).
\end{equation}
The in-context demonstration set for \( t_0 \) is then
\begin{equation}
	\mathcal{D}_{\mathrm{ICL}}(t_0)
	= \{(t_{i_1}, y_{i_1}), \dots, (t_{i_S}, y_{i_S})\},
\end{equation}
where in all experiments  \( S = 10 \), consistent with the setting in
Section~\ref{sec:method}.

The class-level rationales \( \mathcal{R} \) and the MK-GP-selected demonstrations
\( \mathcal{D}_{\mathrm{ICL}}(t_0) \) are concatenated with the query text \( t_0 \)
to form the ICL prompt.
Using \( \oplus \) to denote concatenation, the full prompt supplied to the LLM can
be written as
\begin{equation}
	P(t_0)
	= t_0 \oplus \mathcal{R}
	\oplus (t_{i_1}, y_{i_1}) \oplus \dots \oplus (t_{i_S}, y_{i_S}).
\end{equation}
Given this prompt, the LLM induces a conditional distribution over labels,
\begin{equation}
	p_{\theta}\big(y \mid t_0, \mathcal{R}, \mathcal{D}_{\mathrm{ICL}}(t_0)\big)
	= p_{\theta}\big(y \mid P(t_0)\big),
	\quad y \in \mathcal{Y},
\end{equation}
where \( \theta \) denotes the (frozen) parameters of the LLM.
The predicted label for \( t_0 \) is obtained by a maximum a posteriori decision:
\begin{equation}
	\hat{y}_0
	= \arg\max_{y \in \mathcal{Y}}
	p_{\theta}\big(y \mid t_0, \mathcal{R}, \mathcal{D}_{\mathrm{ICL}}(t_0)\big)
	= \arg\max_{y \in \mathcal{Y}}
	p_{\theta}\big(y \mid P(t_0)\big).
\end{equation}

In this way, the MK-GP-based ICL framework jointly exploits (i) a task-adaptive semantic distance learned
from labeled data and (ii) linguistically rich, label-wise rationales to support
fine-grained sentiment classification under few-shot in-context learning.

\begin{figure}[H]
    \centering
    \begin{promptbox}[{\large\bfseries In-Context Learning}]
        \small\ttfamily
        Analyze the sentiment expressed in the given \textbf{Query Text} toward
        the specified target \{\textit{target}\}.
        The sentiment label must be selected from the following set:
        \{\textit{str(label\_list)}\}.
        Refer to the provided label descriptions and example demonstrations
        to guide your classification.
        
        \vspace{0.5em}
        \textbf{Label Descriptions:} \{Rationales $\mathcal{R}$\} \\
        \textbf{Demonstrations:} \{$(t_1, y_1), \dots, (t_S, y_S)$\}
        
        \vspace{0.5em}
        \textbf{Query Text:} \{\textit{query\_text}\}
    \end{promptbox}

    \caption{Prompt template for the in-context learning (ICL) stage.}
    \label{fig:icl-prompt}
\end{figure}

\section{experiments}
\label{sec:experiments}
\subsection{Experimental Datasets}

To evaluate MK-GP, experiments were conducted on five diverse datasets across various domains and sentiment classification granularities as shown in Table~\ref{table:datasets}:  


\begin{table*}[t]
\centering
\small
\setlength{\tabcolsep}{4pt}
\renewcommand{\arraystretch}{1.15}
\begin{tabularx}{\textwidth}{>{\centering\arraybackslash}m{2.0cm}
                                >{\centering\arraybackslash}m{1.6cm}
                                >{\centering\arraybackslash}m{1.2cm}
                                >{\centering\arraybackslash}m{1.8cm}
                                >{\centering\arraybackslash}X}
\toprule
\textbf{Dataset} & \textbf{Size} & \textbf{Classes} & \textbf{Avg. \#words} & \textbf{Examples} \\
\midrule
\textbf{SST5}\footnotemark[1] & 11,855 & 5 & 17.52 &
{\footnotesize "good film, but very glum.", "formuliac, but fun.", "spiderman rocks"} \\

\textbf{SemEval17}\footnotemark[2] & 20,632 & 5 & 19.78 &
{\footnotesize "see you in milan!", "0"} \\

\textbf{ABSIA}\footnotemark[3] & 4,650 & 7 & 17.50 &
{\footnotesize "food was average but tasty.", "The best in the city!"} \\

\textbf{PR\_Baby}\footnotemark[4] & 25,000 & 5 & 88.62 &
{\footnotesize "Adorable.", "Work perfectly."} \\

\textbf{PR\_Software}\footnotemark[5] & 12,804 & 5 & 184.71 &
{\footnotesize "it works for me.", "gift", "I have not used this yet but appeared to be all there."} \\
\bottomrule
\end{tabularx}
\caption{Summary of experimental datasets with representative examples.}
\label{table:datasets}
\end{table*}

\footnotetext[1]{https://huggingface.co/datasets/SetFit/sst5}
\footnotetext[2]{https://huggingface.co/datasets/midas/semeval2017}
\footnotetext[3]{https://www.iitp.ac.in/\~ai-nlp-ml/resources.html\#ABSIA}
\footnotetext[4]{https://snap.stanford.edu/data/web-Amazon-links.html}
\footnotetext[5]{https://cseweb.ucsd.edu/\~{}jmcauley/datasets/amazon\_v2}

These datasets cover a range of sentiment classification settings, from
sentence-level movie reviews to product and aspect-based opinion texts,
enabling a comprehensive evaluation of MK-GP. For datasets with numeric
labels (ABSIA, SemEval17, PR\_Baby, and PR\_Software), larger label values
correspond to more positive sentiment, while negative values (when present)
indicate negative sentiment.

The original PR\_Baby collection contains 183{,}531 baby-product
reviews.\footnotemark[4] Due to hardware limitations, a stratified, label-balanced subset of 25{,}000 instances (5{,}000 per rating level) is used in all experiments, whereas the other four datasets are used in full.

\begin{figure}[t]
  \centering

  \begin{subfigure}[t]{0.38\linewidth}
    \centering
    \includegraphics[width=\linewidth]{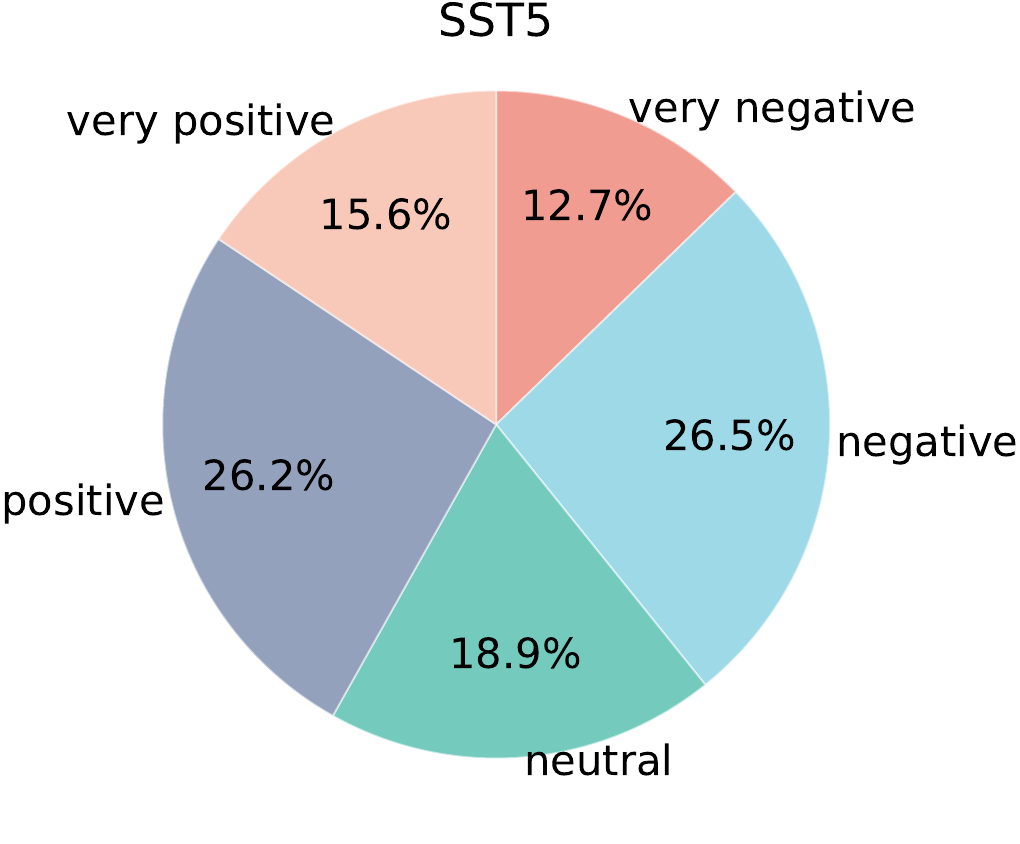}
    \label{fig:pie-sst5}
  \end{subfigure}
  \hspace{2em}
  \begin{subfigure}[t]{0.32\linewidth}
    \centering
    \includegraphics[width=\linewidth]{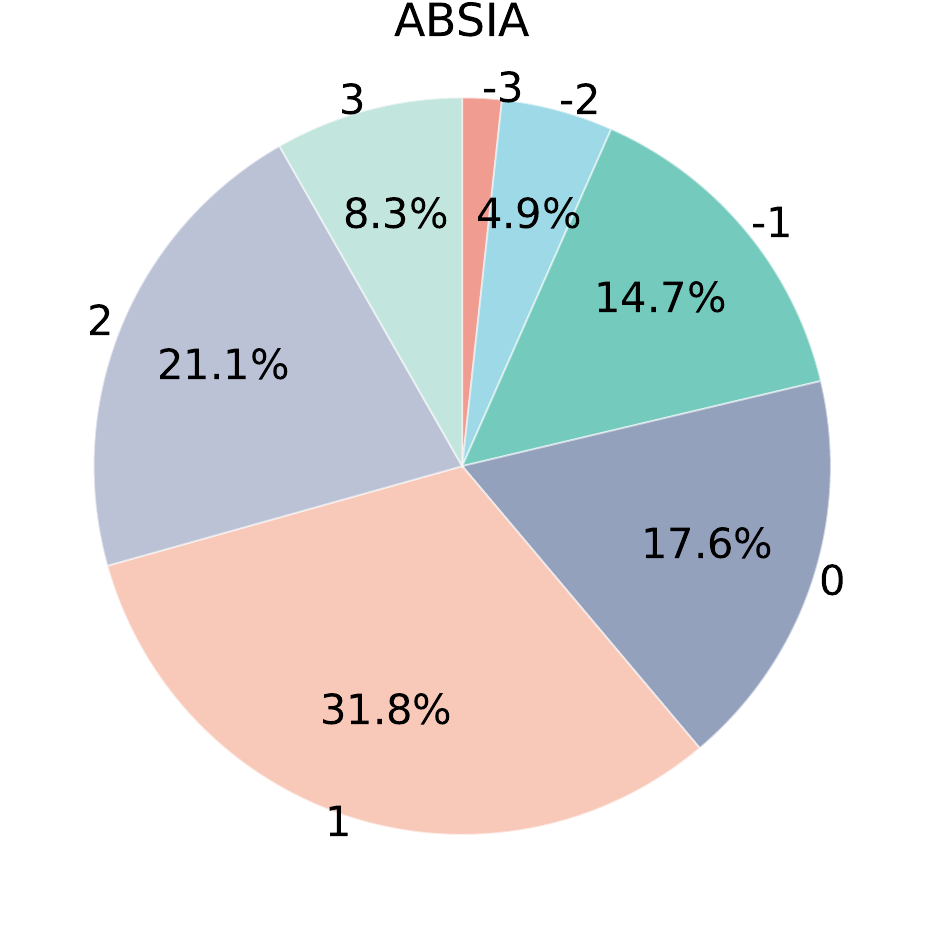}
    \label{fig:pie-absia}
  \end{subfigure}

  \begin{subfigure}[t]{0.32\linewidth}
    \centering
    \includegraphics[width=\linewidth]{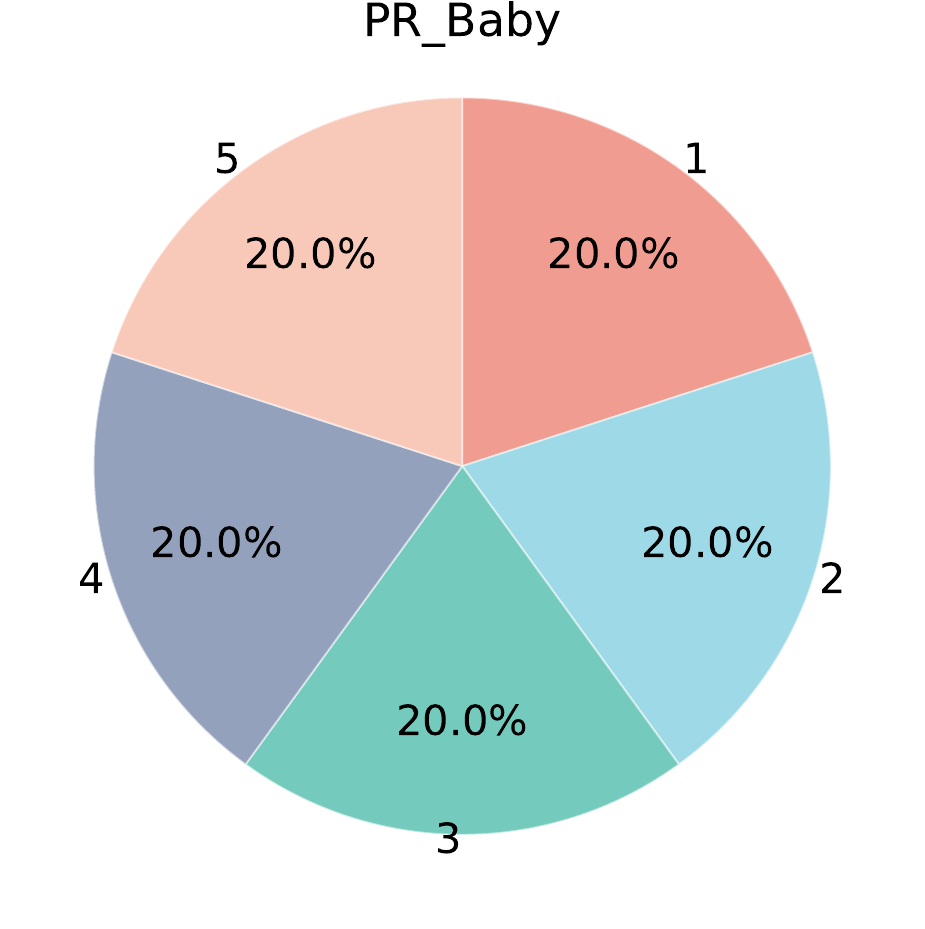}
    \label{fig:pie-prbaby}
  \end{subfigure}
  \hfill
  \begin{subfigure}[t]{0.32\linewidth}
    \centering
    \includegraphics[width=\linewidth]{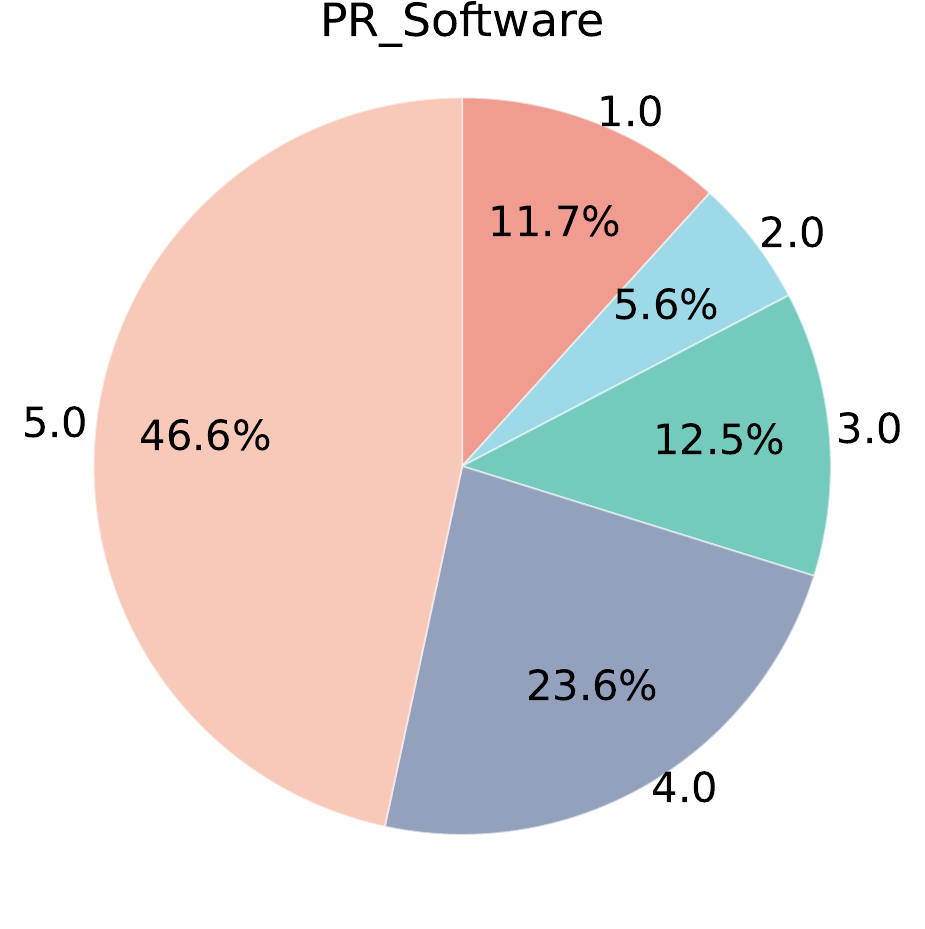}
    \label{fig:pie-prsoft}
  \end{subfigure}
  \hfill
  \begin{subfigure}[t]{0.32\linewidth}
    \centering
    \includegraphics[width=\linewidth]{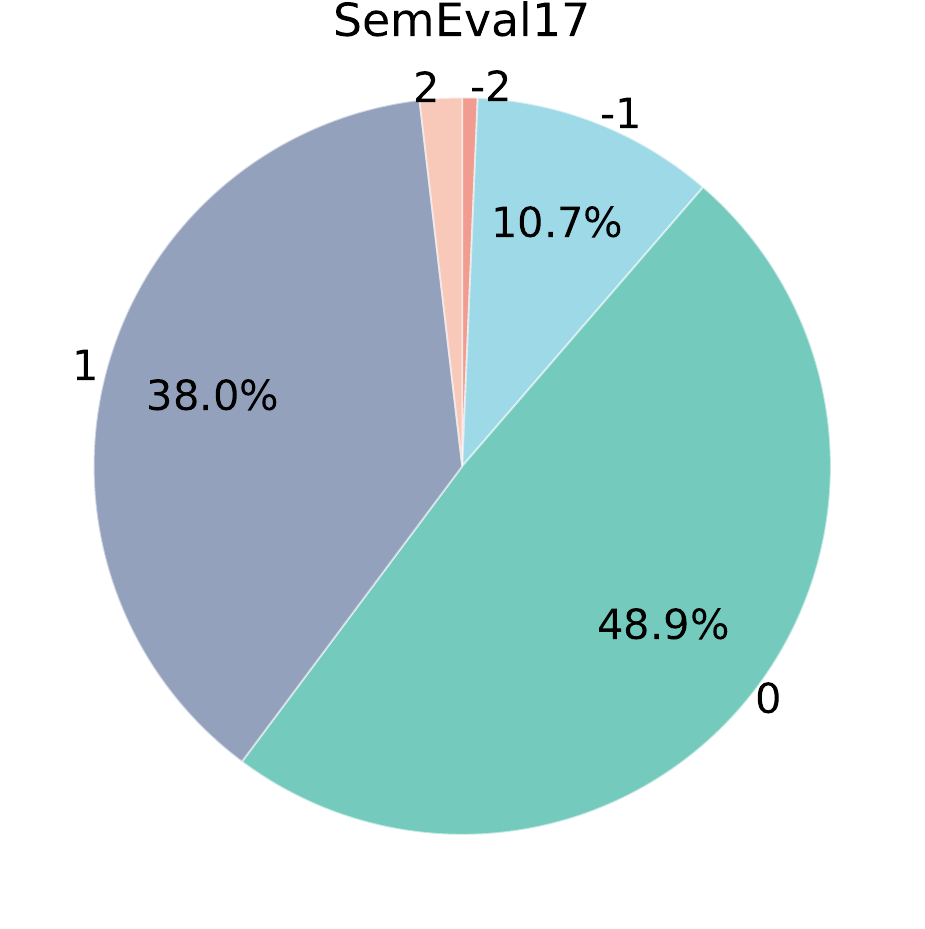}
    \label{fig:pie-semeval}
  \end{subfigure}

  \caption{Label distributions of the five sentiment datasets.}
  \label{fig:label-pies}
\end{figure}

Figure~\ref{fig:label-pies} visualizes the label distributions of the five
datasets. SST5 is moderately imbalanced, ABSIA and PR\_Software are skewed
toward positive labels, the PR\_Baby subset is perfectly balanced, and
SemEval17 is dominated by neutral and mildly positive tweets. This variety of
label granularities and imbalance patterns makes few-shot multi-class
sentiment analysis a non-trivial testbed.

\subsection{Experimental Implementation Details}

In the experiments, 1,000 instances were randomly sampled from each dataset to construct the annotated dataset \( \mathcal{D} \), ensuring a fair evaluation of MK-GP across datasets. This also provided insights into the amount of labeled data required for MCSA tasks, aiding in determining the annotation needed to outperform traditional classifiers trained on large-scale datasets. The balanced Coreset size for generating the classification rationale was set to \( \lambda_\mathcal{B} = 100 \). Taking into account both efficiency and effectiveness, the number of demonstrations was set to \( S = 10 \).

Experiments were conducted using three groups of LLMs: GPT\footnotemark[6], DeepSeek\footnotemark[7], and ERNIE\footnotemark[8]. For each group, the more capable (and expensive) model (GPT-4o, DeepSeek-R1, and ERNIE X1 Turbo) was employed for classification rationale generation, whereas the more cost-efficient variant (GPT-4o-mini, DeepSeek-V3, and ERNIE 4.5 Turbo) was utilized for ICL in MCSA tasks.

\footnotetext[6]{https://openai.com/api/}
\footnotetext[7]{https://www.deepseek.com/}
\footnotetext[8]{https://yiyan.baidu.com/}

The following settings were applied uniformly across all datasets: \( N = 9 \) and \( M = 9 \) were used in the MK-GP model (Equation~(\ref{eq:mk})). The Adam optimizer was adopted with a learning rate of 0.01 over 500 training epochs, and all other optimizer parameters were set to their default values. Optimal hyperparameters were selected via grid search and cross-validation.

Most experiments were conducted on an NVIDIA GeForce RTX 3080 GPU. On average, a single unit of this GPU required 170.86 seconds to complete 500 epochs of Gaussian process training across various datasets. For API-based models, remote inference was employed instead.

\subsection{Comparison Models}

In this study, four In-Context Learning (ICL) methods are compared to evaluate their effectiveness in sentiment classification. The methods selected for comparison are:

\textbf{Random}: In-context examples are selected randomly from the candidate set.

\textbf{BM25}~\cite{robertson2009probabilistic}: The top-$S$ examples are selected using BM25 scoring, which operates on the raw text by considering term frequency and inverse document frequency (TF-IDF) without using embeddings.

\textbf{Cosine Similarity}~\cite{de2022comparing}: The top-$S$ examples are selected based on cosine similarity computed on the embeddings of the candidate set. This method measures the similarity between the text embeddings in a high-dimensional space.

\textbf{MK-GP}: This method uses Multi-Kernel Gaussian Processes (MK-GP) to select the top-$S$ relevant support examples, leveraging a learned, task-aware semantic distance.

All ICL methods select 10 examples ($S = 10$) across five datasets. Additionally, all prompts incorporated classification rationales generated by the same method. Due to the multi-class nature of MCSA and the class imbalance in the experimental data, Accuracy and weighted-average F1 score were used to evaluate the performance of the models.

\subsection{Results and Analysis}

\begin{figure}[H]
    \centering
    \includegraphics[width=0.8\linewidth]{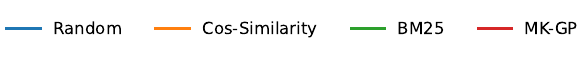}
    
    \vspace{0.8em}
    
    \begin{subfigure}{0.45\linewidth}
        \centering
        \includegraphics[width=\linewidth]{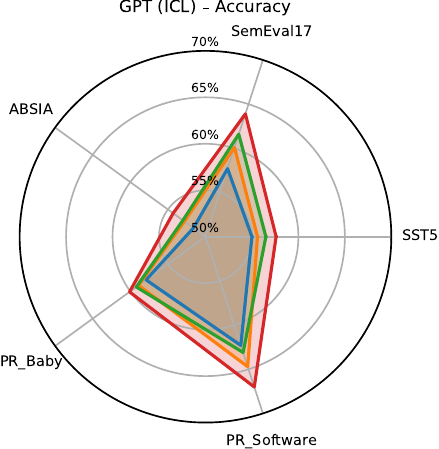}
    \end{subfigure}
    \hfill
    \begin{subfigure}{0.45\linewidth}
        \centering
        \includegraphics[width=\linewidth]{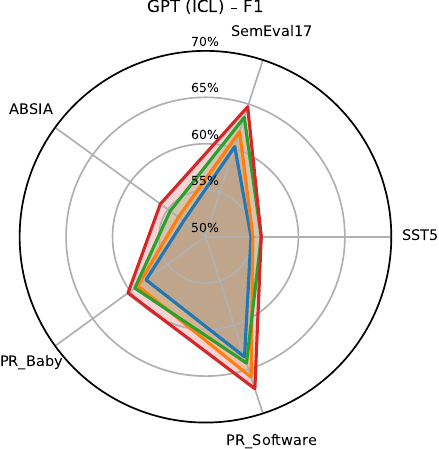}
    \end{subfigure}
    
    \vspace{0.8em}
    
    \begin{subfigure}{0.45\linewidth}
        \centering
        \includegraphics[width=\linewidth]{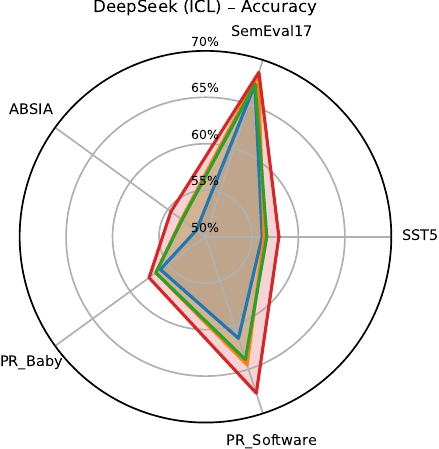}
    \end{subfigure}
    \hfill
    \begin{subfigure}{0.45\linewidth}
        \centering
        \includegraphics[width=\linewidth]{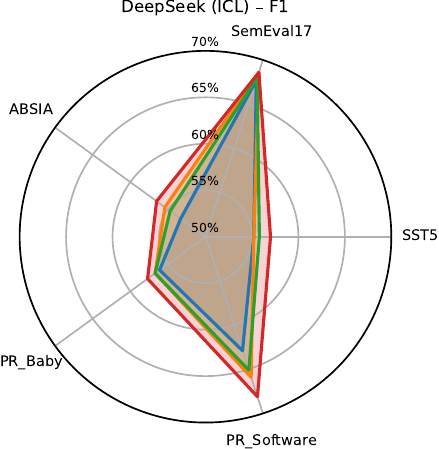}
    \end{subfigure}
    
    \vspace{0.8em}
    
    \begin{subfigure}{0.45\linewidth}
        \centering
        \includegraphics[width=\linewidth]{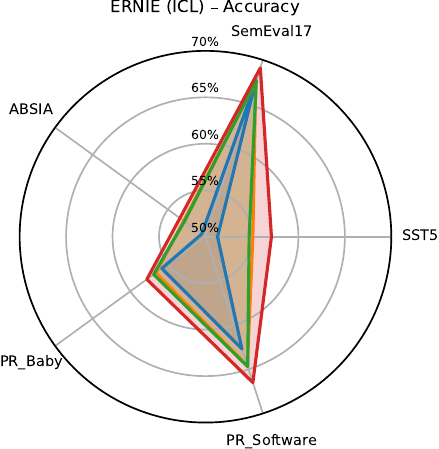}
    \end{subfigure}
    \hfill
    \begin{subfigure}{0.45\linewidth}
        \centering
        \includegraphics[width=\linewidth]{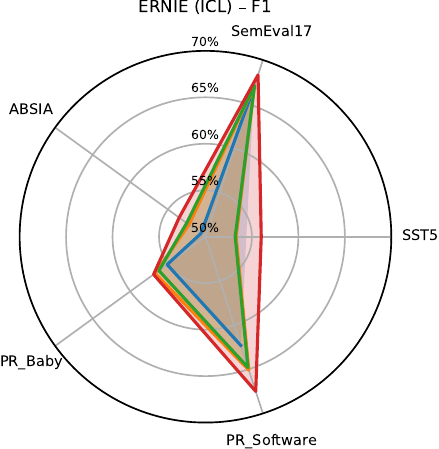}
    \end{subfigure}
    
    \caption{Radar plots of four ICL selection strategies across five datasets under three backbone LLM families.}
    \label{fig:icl-radar-all}
\end{figure}

The overall results of the four ICL selection strategies across the five datasets and three LLM backbones are summarized in Figure~\ref{fig:icl-radar-all}. Across all three LLM families, MK-GP consistently attained the best scores on every dataset in terms of both Accuracy and F1. The methods for comparison exhibited broadly similar behavior across datasets: Random selection was consistently weakest, while Cosine Similarity and BM25 offered modest and closely clustered gains. By contrast, MK-GP shifted the curves outward in a stable way across movie reviews, tweets, aspect-based restaurant reviews, and product reviews. This suggested that replacing hand-designed similarity measures (cosine or BM25) with a learned, task-aware MK-GP semantic metric yielded consistent, architecture-agnostic improvements for ICL-based multi-class sentiment analysis.

\subsection{Ablation Study}
\label{sec:ablation}

An ablation study was conducted to isolate the role of the kernel design in MK-GP, focusing on how different covariance structures affected semantic similarity estimation. Three variants were compared: \textbf{using only the non-stationary kernel}, \textbf{using only the stationary kernel}, and \textbf{using the full Mat\'ern+polynomial combined kernel}. All remaining settings were held constant, including the labeled pool size, the balanced coreset for rationale generation, and the number of retrieved demonstrations ($S=10$).

\begin{figure*}[t]
    \centering
    \includegraphics[width=1.5\columnwidth]{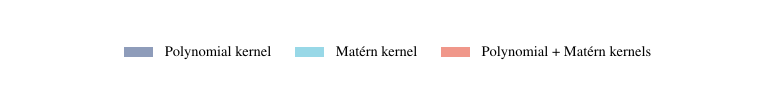}\par
    
    \includegraphics[width=0.31\textwidth]{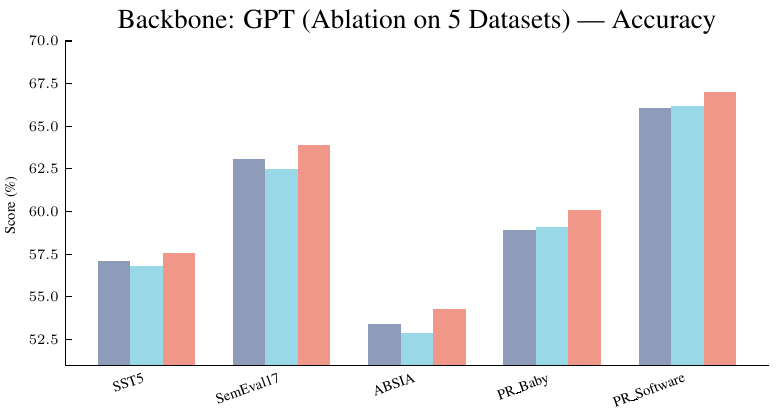}\hfill
    \includegraphics[width=0.31\textwidth]{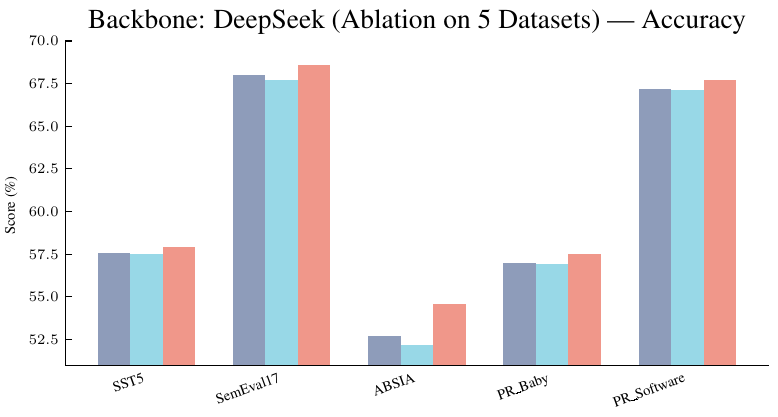}\hfill
    \includegraphics[width=0.31\textwidth]{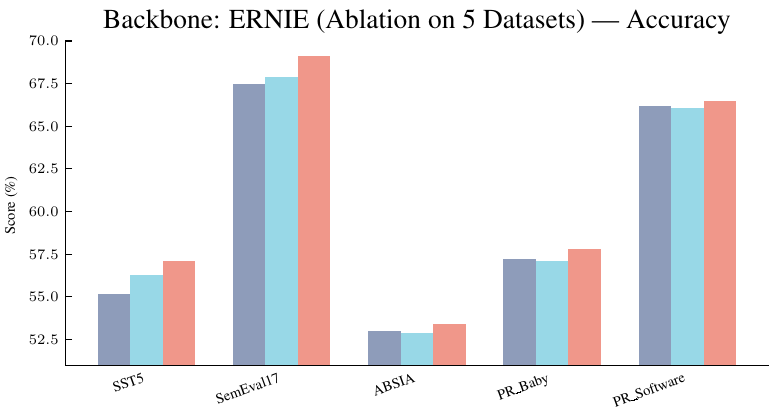}\par

    \includegraphics[width=0.31\textwidth]{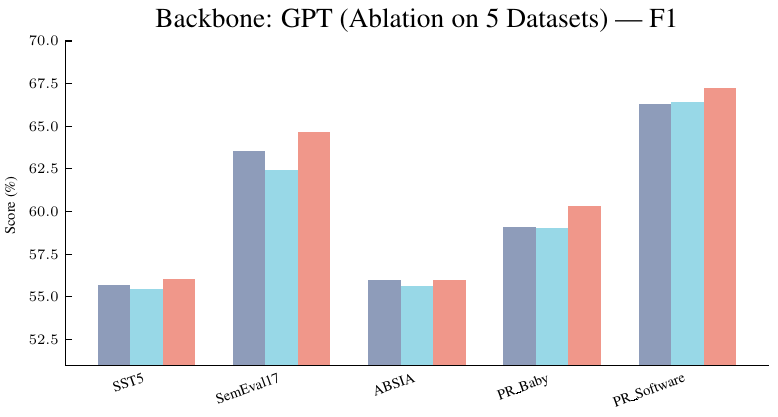}\hfill
    \includegraphics[width=0.31\textwidth]{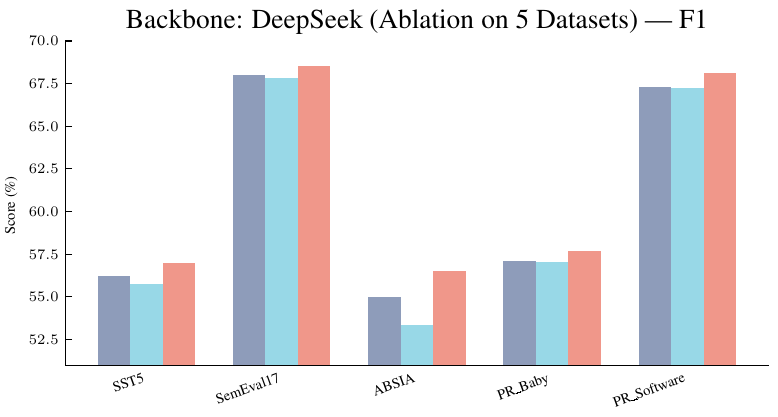}\hfill
    \includegraphics[width=0.31\textwidth]{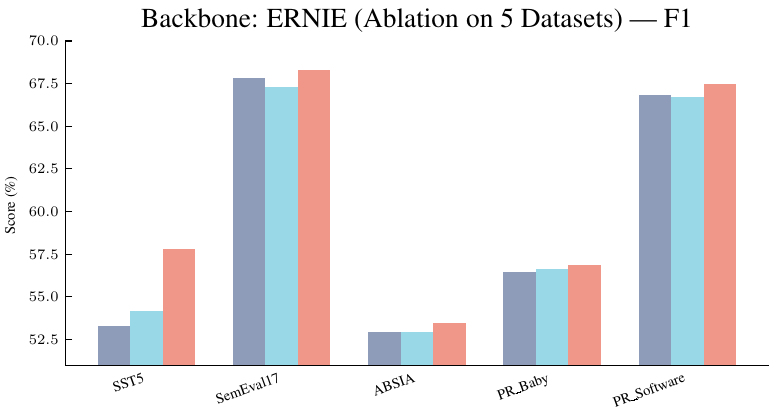}\par

    \caption{Ablation results on five datasets. Accuracy and F1 scores for three backbones: GPT, DeepSeek, and ERNIE.}
    \label{fig:ablation_kernels_5ds_acc_f1}
\end{figure*}

Figure~\ref{fig:ablation_kernels_5ds_acc_f1} reports Accuracy and weighted-F1 across five datasets and three LLM backbones. The single-kernel variants were competitive but exhibited dataset- and backbone-dependent variability. This pattern supported the motivation for combining kernels: The Mat\'ern kernel was well-suited to characterize the {stationary} structure in the embedding space, whereas the polynomial kernel could capture {non-stationary} variations by modeling higher-order feature interactions. Their combination yielded a more faithful task-aware semantic distance, leading to more reliable demonstration selection in ICL.

\section{Conclusions}
\label{sec:conclusion}

This paper has approached in-context multi-class sentiment analysis from a task-aware metric learning perspective. Instead of relying on fixed, task-agnostic similarities such as BM25 or cosine distance, a MK-GP is used to learn a semantic kernel that is explicitly aligned with the label geometry in a reproducing kernel Hilbert space. On top of this learned metric, the proposed MK-GP framework combines class-level rationales induced from a label-balanced coreset with similarity-based demonstration selection to construct more informative prompts for in-context learning.Extensive experiments on five fine-grained sentiment benchmarks and three families of large language models validate the effectiveness of the proposed method.

Although this work focuses on multi-class sentiment analysis as a primary application, the underlying idea of learning task-aware semantic distances is more general. Future directions include applying the learned kernels to RAG, extending the metric to multimodal settings for cross-modal alignment, and exploring more scalable GP approximations and tighter integration with LLM training and prompting strategies.

\bibliography{custom}

\appendix


\end{document}